\documentclass[nohyperref]{article}

\usepackage{microtype}
\usepackage{graphicx}
\usepackage{subfigure}
\usepackage{booktabs} 
\usepackage{floatrow}
\usepackage{acronym}
\newacro{TM}{Tsetlin Machine}
\newacro{TA}{Tsetlin Automata}
\usepackage{hyperref}
\usepackage{multicol}
\usepackage{multirow}

 \usepackage[accepted]{icml2022}

\usepackage{amsmath}
\usepackage{amssymb}
\usepackage{mathtools}
\usepackage{amsthm}
\DeclareMathOperator*{\argmax}{arg\,max}

\usepackage[capitalize,noabbrev]{cleveref}

\theoremstyle{plain}

\theoremstyle{definition}

\theoremstyle{remark}

\usepackage[textsize=tiny]{todonotes}

\icmltitlerunning{Tsetlin Machine for Solving Contextual Bandit Problems}

\begin{document}

\twocolumn[
\icmltitle{Tsetlin Machine for Solving Contextual Bandit Problems}




\begin{icmlauthorlist}
\icmlauthor{Raihan Seraj}{yyy}
\icmlauthor{Jivitesh Sharma}{zzz}
\icmlauthor{Ole-Christoffer Granmo}{zzz}
\end{icmlauthorlist}

\icmlaffiliation{yyy}{Department of Electrical \& Computer Engineering, McGill University, Montreal, Canada}
\icmlaffiliation{zzz}{Center for Artificial Intelligence Research, University of Agder, Kristiansand, Norway}

\icmlcorrespondingauthor{Raihan Seraj}{raihan.seraj@mail.mcgill.ca}

\icmlkeywords{Contextual Bandits, Tsetlin Machine, Online Learning}

\vskip 0.3in
]



\printAffiliationsAndNotice{}  

\begin{abstract}
This paper introduces an interpretable contextual bandit algorithm using Tsetlin Machines, which solves complex pattern recognition tasks using propositional logic. The proposed bandit learning algorithm relies on straightforward bit manipulation, thus simplifying computation and interpretation. We then present a mechanism for performing Thompson sampling with Tsetlin Machine, given its non-parametric nature. Our empirical analysis shows that Tsetlin Machine as a base contextual bandit learner outperforms other popular base learners on eight out of nine datasets. We further analyze the interpretability of our learner, investigating how arms are selected based on propositional expressions that model the context\footnote{The code is available online on: \href{https://github.com/Raihan-Seraj/Tsetlin-Machine-for-Solving-Contextual-Bandit-Problems}{github}}.


\end{abstract}

\section{Introduction }
Contextual bandits play a fundamental role in many applications involving sequential decision making, ranging from personalized recommendations of movies or products~\cite{li2010contextual} to designing effective treatment allocation strategies in clinical trials~\cite{durand2018contextual,bouneffouf2020survey}. Algorithms for contextual bandits have additionally gained significant interest because of their theoretical elegance. In brief, a decision-maker selects one of multiple bandit arms over a sequence of rounds, taking into account an observed context. Each round, the arm chosen elicits feedback in the form of a reward signal associated with the success of selecting that arm (such as a user purchasing a recommended product). The contextual bandit problem is particularly intriguing because the decision-maker must maximize the expected reward in as few rounds as possible, trading exploitation against exploration to identify the optimal arm.


In this paper, we recast the \ac{TM} \cite{granmo2018tsetlin} as a contextual bandit algorithm and study the resulting scheme empirically. \ac{TM} is a recent approach to pattern recognition that employs a team of non-contextual bandit algorithms, in the form of Tsetlin automata \cite{Tsetlin1961}, to learn patterns expressed in propositional logic. \acp{TM} have been shown to obtain competitive accuracy, memory footprint, and learning speed on several benchmark datasets~\cite{abeyrathna2021massively}. They have been particularly successful in natural language processing, including explainable aspect-based sentiment analysis~\cite{yadav2021human}. Being based on finite state automata, they further support Markov chain-based convergence analysis \cite{zhang2021convergence}. Leveraging the non-linear pattern recognition capability of \acp{TM}, our proposed scheme thus addresses the contextual bandit problem using a team of non-contextual bandit algorithms.

Although existing algorithms for contextual bandits provide theoretical guarantees, they are either difficult to interpret or make assumptions that limit their usability in real-world settings. Motivated by these limitations, our work contributes as follows:
\begin{itemize}
    \item We investigate \ac{TM} as a base learner for contextual bandit problems and empirically demonstrate its effectiveness compared to other popular base learners. 
    \item We propose how the Thompson sampling scheme can be leveraged by \acp{TM} through bootstrapping \cite{practicalcbdt}.
    \item We provide interpretability analysis that shows how algorithms using \acp{TM} select arms based on propositional expressions of context features. 
\end{itemize}

To the best of our knowledge, this is the first reported work on learning arm selection strategies for contextual bandits expressed in propositional logic.
The paper is organized as follows. In Section~\ref{sec:lit_overview}, we provide a literature overview on contextual bandits and different algorithms for solving them. Section~\ref{sec:problem_formulation} and \ref{sec:TM} present the problem formulation and introduces  \ac{TM}, respectively. In Section~\ref{sec:TM_cb}, we show how \ac{TM} can be used as a base learner for contextual bandits. The numerical results and the interpretability analysis are then presented in Section~\ref{sec:numerical} and~\ref{sec:interpretability} respectively. We conclude the paper and provide directions for future work in Section~\ref{sec:conclusion}.

\section{Literature Overview}
\label{sec:lit_overview}
Existing works on contextual bandits can be broadly categorized into two categories~\cite{practicalcbdt}. The first category includes algorithms that focus on exploration problems while the second category pertains to using different base learners for contextual bandits. Literature focusing on the exploration problems provide theoretical guarantees on regret bounds, however, the bounds are often tied with very strong underlying assumptions. In~\cite{chu2011contextual}, for instance, the authors provide a regret bound that holds with probability $1-\delta$ and is of the order $\mathcal O\Bigl(\sqrt{Td\ln^3(KT\ln(T)/\delta}\Bigr)$ for $d$ dimensional context vector with $T$ rounds and $K$ actions. The underlying assumption for obtaining such regret bounds is that the payoff (or reward) is a linear function of the context features. Similar assumptions underlie the work in~\cite{agrawal2013thompson}, where the authors present Thompson sampling for contextual bandits with linear payoffs. 

For the second category, different base learners for contextual bandits have been investigated. In~\cite{allesiardo2014neural}, the authors employ neural networks to model the value of rewards given the contexts. Subsequently deep learning for contextual bandits have also been explored in~\cite{zhou2020neural,shen2018interactive,ismath2021deep}. In~\cite{feraud2016random}, the authors use random forest as a base learner for contextual bandits. The proposed learner is optimal up to a logarithmic factor where the computational cost of the algorithm is linear with respect to the time horizon. In~\cite{practicalcbdt}, the authors use decision tree learners for contextual bandits, and then propose Thompson sampling for such non-parametric learners. 

Our work pertains to the second category,  where we use \ac{TM} as a base learner for contextual bandits. 
Our \ac{TM} learner supports incremental training with streaming data. In contrast to popular baseline learners such as artificial neural networks, both the learned arm-context model and the process of learning are easy to follow and explain. The interpretability is attained using propositional functions of context features used by \ac{TM} for arm selection.

\section{Problem Formulation}
\label{sec:problem_formulation}
We consider an online stochastic contextual bandit setup where at time $t$, context-reward pair denoted by $(s_t,r_t)$ is sampled independently from past data distribution $\mathcal D$. Here $s_t\in\mathcal S$ represents an $M$ dimensional context vector and $r_t=((r_t(1),\dots,r_t(K))\in \{0,1\}^K$ is the reward vector for $K$ possible actions. The learner chooses an arm $u_t\in\{1,\dots,K\}$ after observing the context $s_t$ and receives a reward $r_t(u_t)$ for the chosen arm. The objective of the learner is to perform a sequence of actions in order to minimize the cumulative expected regret given by 
\begin{equation}
\label{eq:regret}
    Regret = \mathbb{E}\Bigl[\sum_{t=1}^T\Bigl(r_t(\pi^*(s_t))-r_t(u_t)\Bigr)\Bigr].
\end{equation}
Here, $\pi^*=\displaystyle\argmax_{\pi\in\Pi}\mathbb{E}_{(s,r)\sim \mathcal D}[r(\pi(s))]$ where $\Pi$ denotes the set of large (possiblly infinite) policies and $\pi:\mathcal S\mapsto \{1,\dots,K\}$. For our analysis we  consider maximizing the expected total reward which is equivalent to minimizing regret.






\section{Tsetlin Machine}
\label{sec:TM}
\begin{figure}[!thb]
  \begin{center}
  \includegraphics[width=\linewidth]{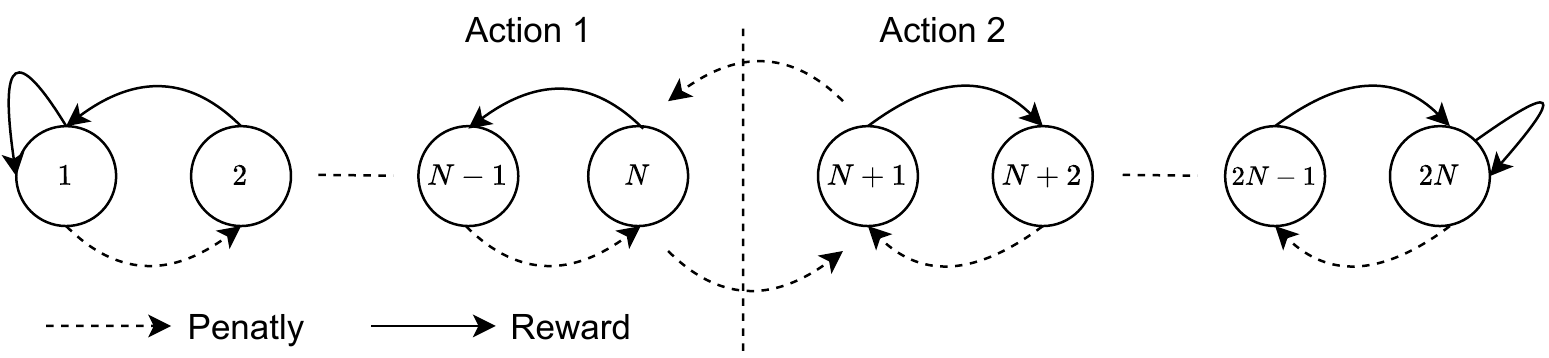}
  \caption{A two-action Tsetlin Automaton with $2N$ states.}\label{figTA}
  \end{center}
\end{figure}

\begin{figure}[ht]
\begin{center}
\centerline{\includegraphics[width=\columnwidth]{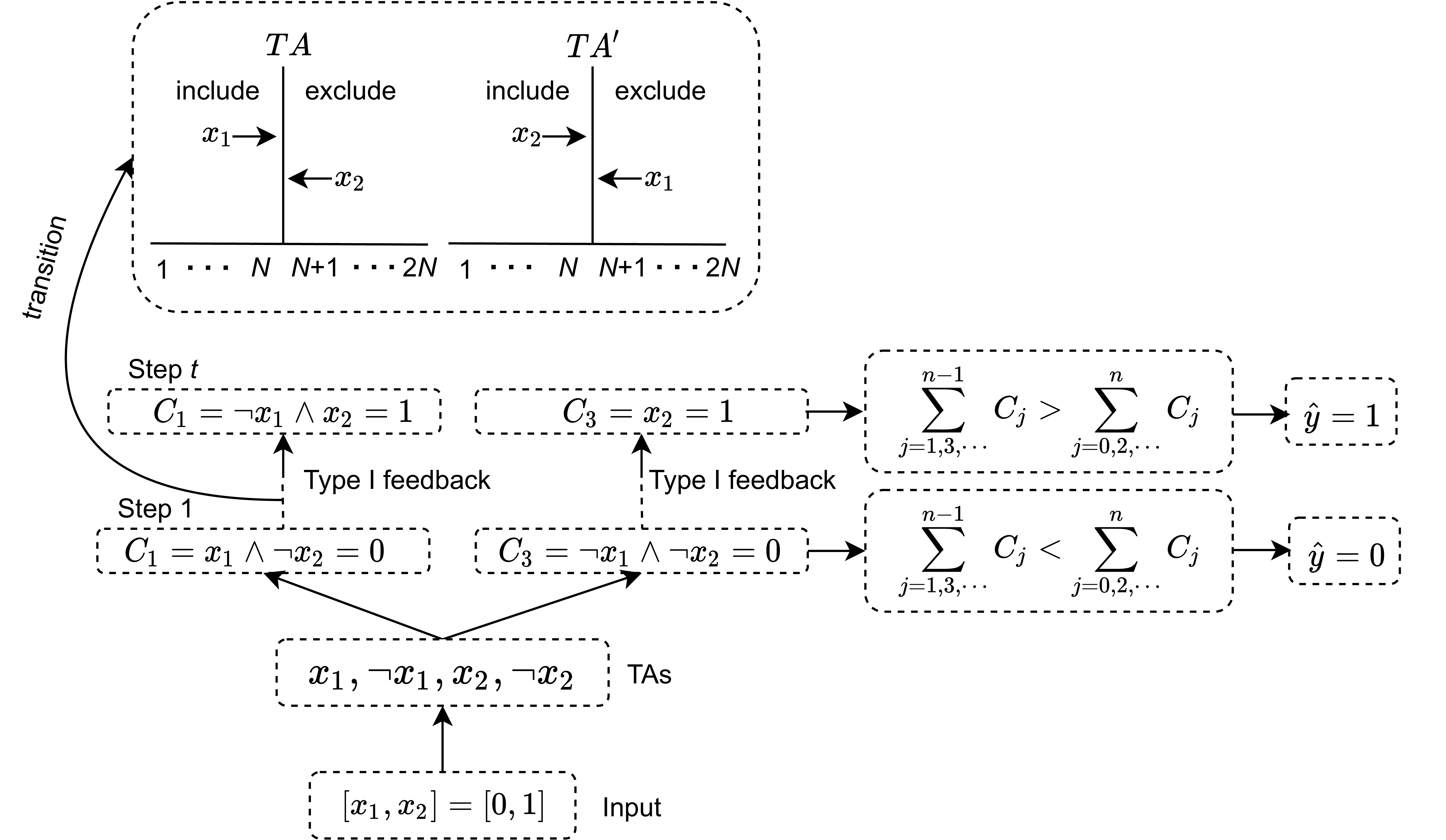}}
\caption{TM learning dynamics for an XOR-gate training sample, with input ($x_1=0, x_2=1$) and output target $y=1$.}
\label{figure:tm_architecture_basic}
\end{center}
\end{figure}

\paragraph{Structure.} A \ac{TM} in its simplest form takes a feature vector $\mathbf{x} = [x_1, x_2, \ldots, x_o] \in \{0,1\}^o$ of $o$ propositional values as input and assigns the vector a class $\hat{y} \in \{0,1\}$. To minimize classification error, the TM produces $n$ self-contained patterns. In brief, the input vector $\mathbf{x}$ provides the literal set $L = \{l_1, l_2, \ldots, l_{2o}\} = \{x_1, x_2, \ldots, x_{o}, \lnot x_1, \lnot x_2, \ldots, \lnot x_o\}$, consisting of the input features and their negations. By selecting subsets $L_j \subseteq L$ of the literals, the TM can build arbitrarily complex patterns, ANDing the selected literals to form conjunctive clauses:
\begin{equation}
C_j(\mathbf{x})= \bigwedge_{l_k \in L_j} l_k.
\end{equation}
Above, $j \in \{1, 2, \ldots, n\}$ refers to a particular clause $C_j$ and $k \in \{1, 2, \ldots, 2o\}$ refers to a particular literal $l_k$. As an example, the clause $C_j(\mathbf{x}) = x_1 \land \lnot x_2$ consists of the literals $L_j = \{x_1, \lnot x_2\}$ and evaluates to $1$ when $x_1=1$ and $x_2=0$. 

The TM assigns one \ac{TA} per literal $l_k$ per clause $C_j$ to build the clauses. The \ac{TA} assigned to literal $l_k$ of clause $C_j$ decides whether $l_k$ is \emph{Excluded} or \emph{Included} in $C_j$. Figure~\ref{figTA} depicts a two-action \ac{TA} with $2N$ states.  For states $1$ to $N$, the \ac{TA} performs action \emph{Exclude} (Action 1), while for states $N + 1$ to $2N$ it performs action \emph{Include} (Action 2). As feedback to the action performed, the environment responds with either a Reward or a Penalty. If the TA receives a Reward, it moves deeper into the side of the action. If it receives a Penalty, it moves towards the middle and eventually switches action.

With $n$ clauses and $2o$ literals, we get $n\times2o$ \ac{TA}s. We organize the states of these in a $n\times2o$ matrix $A = [a_k^j] \in \{1, 2, \ldots, 2N\}^{n\times2o}$. We will use the function $g(\cdot)$ to map the automaton state $a_k^j$ to Action $0$ (\emph{Exclude}) for states $1$ to $N$ and to Action $1$ (\emph{Include}) for states $N+1$ to~$2N$: $g(a_k^j) = a_k^j > N$.

We can connect the states $a_k^j$ of the \ac{TA}s assigned to clause $C_j$ with its composition as follows:
\begin{equation}
C_j(\mathbf{x}) = \bigwedge_{l_k \in L_j} l_k = \bigwedge_{k=1}^{2o} \left[g(a_k^j) \Rightarrow l_k\right].
\end{equation}
Here, $l_k$ is one of the literals and $a_k^j$ is the state of its \ac{TA} in clause $C_j$. The logical \emph{imply} operator~$\Rightarrow$ implements the \emph{Exclude}/\emph{Include} action. That is, the 
\emph{imply} operator is always $1$ if $g(a_k^j)=0$ (\emph{Exclude}), while if $g(a_k^j)=1$ (\emph{Include}) the truth value is decided by the truth value of the literal.

\paragraph{Classification.} Classification is performed as a majority vote. The odd-numbered half of the clauses vote for class $\hat{y} = 0$ and the even-numbered half vote for $\hat{y} = 1$:
\begin{equation}
    \hat{y} = 0 \le \sum_{j=1,3,\ldots}^{n-1} \bigwedge_{k=1}^{2o} \left[g(a_k^j) \Rightarrow l_k\right] - \sum_{j=2,4,\ldots}^{n} \bigwedge_{k=1}^{2o} \left[g(a_k^j) \Rightarrow l_k\right]. \label{eqn:prediction}
\end{equation}
As such, the odd-numbered clauses have positive polarity, while the even-numbered ones have negative polarity. As an example, consider the input vector $\mathbf{x} = [0, 1]$ in the lower part of Figure \ref{figure:tm_architecture_basic}. The figure depicts two clauses of positive polarity, $C_1(\mathbf{x}) = x_1 \land \lnot x_2$ and $C_3(\mathbf{x}) = \lnot x_1 \land \lnot x_2$ (the negative polarity clauses are not shown). Both of the clauses evaluate to zero, leading to class prediction $\hat{y} = 0$.

\begin{table}[t]
\centering
\vskip 0.15in
\begin{center}
\begin{small}
\begin{sc}
\begin{tabular}{l|l|l|l}
    \hline
    \multirow{2}{*}{Input}&Clause & \ \ \ \ \ \ \ 1 & \ \ \ \ \ \ \ 0 \\
    &{Literal} &\ \ 1 \ \ \ \ \ \ 0 &\ \ 1 \ \ \ \ \ \ 0 \\
    \hline
    \multirow{2}{*}{Include Literal}&P(Reward)&$\frac{s-1}{s}$\ \ \ NA & \ \ 0 \ \ \ \ \ \ 0\\ [1mm]
    &P(Inaction)&$\ \ \frac{1}{s}$\ \ \ \ \ NA &$\frac{s-1}{s}$ \ $\frac{s-1}{s}$ \\ [1mm]
    &P(Penalty)& \ \ 0 \ \ \ \ \ NA& $\ \ \frac{1}{s}$ \ \ \ \ \  $\frac{1}{s}$ \\ [1mm]
    \hline
    \multirow{2}{*}{Exclude Literal}&P(Reward)& \ \ 0 \ \ \ \ \ \ $\frac{1}{s}$ & $\ \ \frac{1}{s}$ \ \ \ \ \
    $\frac{1}{s}$ \\ [1mm]
    &P(Inaction)&$ \ \ \frac{1}{s}$\ \ \ \ $\frac{s-1}{s}$  &$\frac{s-1}{s}$ \ $\frac{s-1}{s}$ \\ [1mm]
    &P(Penalty)&$\frac{s-1}{s}$ \ \ \ \ 0& \ \ 0 \ \ \ \ \ \ 0 \\ [1mm]
    \hline
\end{tabular}
\end{sc}
\end{small}
\end{center}
\caption{Type I Feedback}
\label{table:type_i}
\end{table}

\begin{table}[t]
\centering
\vskip 0.15in
\begin{center}
\begin{small}
\begin{sc}
\begin{tabular}{l|l|l|l}
    \hline
    \multirow{2}{*}{Input}&Clause & \ \ \ \ \ \ \ 1 & \ \ \ \ \ \ \ 0 \\
    &{Literal} &\ \ 1 \ \ \ \ \ \ 0 &\ \ 1 \ \ \ \ \ \ 0 \\
    \hline
    \multirow{2}{*}{Include Literal}&P(Reward)&\ \ 0 \ \ \ NA & \ \ 0 \ \ \ \ \ \ 0\\[1mm]
    &P(Inaction)&1.0 \ \  NA &  1.0 \ \ \ 1.0 \\[1mm]
    &P(Penalty)&\ \ 0 \ \ \ NA & \ \ 0 \ \ \ \ \ \ 0\\[1mm]
    \hline
    \multirow{2}{*}{Exclude Literal}&P(Reward)&\ \ 0 \ \ \ \ 0 & \ \ 0 \ \ \ \ \ \ 0\\[1mm]
    &P(Inaction)&1.0 \ \ \ 0 &  1.0 \ \ \ 1.0 \\[1mm]
    &P(Penalty)&\ \ 0 \ \  1.0 & \ \ 0 \ \ \ \ \ \ 0\\[1mm]
    \hline
\end{tabular}
\end{sc}
\end{small}
\end{center}
\caption{Type II Feedback}
\label{table:type_ii}
\end{table}

\paragraph{Learning.} The upper part of Figure \ref{figure:tm_architecture_basic} illustrates learning. A TM learns online, processing one training example $(\mathbf{x}, y)$ at a time. Based on $(\mathbf{x}, y)$, the TM rewards and penalizes its \ac{TA}s, which amounts to incrementing and decrementing their states. There are two kinds of feedback: Type I Feedback produces frequent patterns and Type II Feedback increases the discrimination power of the patterns.

Type I feedback is given stochastically to clauses with positive polarity when $y=1$  and to clauses with negative polarity when $y=0$. Conversely, Type II Feedback is given stochastically to clauses with positive polarity when $y=0$ and to clauses with negative polarity when $y=1$. The probability of a clause being updated is based on the vote sum $v$: $v = \sum_{j=1,3,\ldots}^{n-1} \bigwedge_{k=1}^{2o} \left[g(a_k^j) \Rightarrow l_k\right] - \sum_{j=2,4,\ldots}^{n} \bigwedge_{k=1}^{2o} \left[g(a_k^j) \Rightarrow l_k\right]$. The voting error is calculated as:
\begin{equation}
\epsilon = \begin{cases}
T-v& y=1\\
T+v& y=0.
\end{cases}
\end{equation}
Here, $T$ is a user-configurable voting margin yielding an ensemble effect. The probability of updating each clause is $P(\mathrm{Feedback}) = \frac{\epsilon}{2T}$.

After random sampling from $P(\mathrm{Feedback})$ has decided which clauses to update, the following \ac{TA} state updates can be formulated as matrix additions, subdividing Type I Feedback into feedback Type Ia and Type Ib:
\begin{equation}
    A^*_{t+1} = A_t + F^{\mathit{II}} + F^{Ia} - F^{Ib}.
    \label{eqn:learning_step_1}
\end{equation}
Here, $A_t = [a^j_k] \in \{1, 2, \ldots, 2N\}^{n \times 2o}$ contains the states of the \ac{TA}s at time step $t$ and $A^*_{t+1}$ contains the updated state for time step $t+1$ (before clipping). The matrices $F^{\mathit{Ia}} \in \{0,1\}^{n \times 2o}$ and $F^{\mathit{Ib}} \in \{0,1\}^{n \times 2o}$ contains Type I Feedback. A zero-element means no feedback and a one-element means feedback. As shown in Table \ref{table:type_i}, two rules govern Type I feedback:
\begin{itemize}
    \item \textbf{Type Ia Feedback} is given with probability $\frac{s-1}{s}$ whenever both clause and literal are $1$-valued.\footnote{Note that the probability $\frac{s-1}{s}$ is replaced by $1$ when boosting true positives.} It penalizes \emph{Exclude} actions and rewards \emph{Include} actions. The purpose is to remember and refine the patterns manifested in the current input $\mathbf{x}$. This is achieved by increasing selected \ac{TA} states. The user-configurable parameter $s$ controls pattern frequency, i.e., a higher $s$ produces less frequent patterns.
    \item \textbf{Type Ib Feedback} is given with probability $\frac{1}{s}$ whenever either clause or literal is $0$-valued. This feedback rewards \emph{Exclude} actions and penalizes \emph{Include} actions to coarsen patterns, combating overfitting. Thus, the selected \ac{TA} states are decreased.
\end{itemize}

The matrix $F^{\mathit{II}} \in \{0, 1\}^{n \times 2o}$ contains Type II Feedback to the \ac{TA}s, given per Table \ref{table:type_ii}.
\begin{itemize}
\item \textbf{Type II Feedback} penalizes \emph{Exclude} actions to make the clauses more discriminative, combating false positives. That is, if the literal is $0$-valued and the clause is $1$-valued, \ac{TA} states below $N+1$ are increased. Eventually the clause becomes $0$-valued for that particular input, upon inclusion of the $0$-valued literal.
\end{itemize}

The final updating step for training example  $(\mathbf{x}, y)$ is to clip the state values to make sure that they stay within value $1$ and $2N$:
\begin{equation}
    A_{t+1} = \mathit{clip}\left(A^*_{t+1}, 1, 2N\right). \label{eqn:learning_step_2}
\end{equation}

For example, both of the clauses in Figure~\ref{figure:tm_architecture_basic} receives Type I Feedback over several training examples, making them resemble the input associated with $y=1$.

\section{Contextual Bandits with Tsetlin Machines}
\label{sec:TM_cb}
We use 
\ac{TM} as a contextual bandit learner that learns a  mapping from context to actions incrementally with streaming data. Since both inputs, patterns and outputs of TM are represented as bits~\cite{granmo2018tsetlin}, 
each $M$ dimensional context $s_t$ is binarized with appropriate number of bits. This results in a $B$ dimensional binarized context with $B\geq M$ which is  fed to each \ac{TM} learner. We outline appropriate choices of bits for binarization for different datasets in the next section. The \ac{TM} learner, being non-parametric, has the advantage that it makes few or no assumption about the underlying functions to be learned; hence they are adaptive and exhibit high degree of flexibility. Another popular non-parametric learner for contextual bandits is the decision tree, which has been thoroughly studied in the literature~\cite{practicalcbdt,soemers2018adapting,feraud2016random}. The \ac{TM}, in contrast, learns a linear combination of conjunctive clauses in propositional logic by producing decision rules similar to the branches in decision trees~\cite{abeyrathna2021extending}, with the added advantage of being memory efficient, computationally simple and not having a tendency to overfit the training data.
We consider two contextual bandit learning algorithms using TM: (i) TM with epsilon greedy arm selection; and (ii) TM with Thompson sampling. 
\par
\emph{Epsilon greedy \ac{TM}:} The exploration-exploitation trade-off is a fundamental problem in learning to make decisions under uncertainty. In a multi-armed bandit setting, the $\varepsilon-$greedy algorithm is one of the simplest ones.  The learner either chooses the empirically best arm with probability  $(1-\varepsilon)$ (exploitation) or a random arm with probability $\varepsilon$ (exploration). In the contextual bandit setting, given a set of contexts, the learner chooses to select the current empirically best arm with some high probability, maximizing immediate rewards. Otherwise, it selects a randomized arm with the hope of improving future rewards given the context. Variations of epsilon greedy algorithms have been explored in the literature, including decaying the $\varepsilon$ parameter and eventually dropping the probability of choosing a random arm to zero. 
In our setting, we consider a fixed small value of $\varepsilon$ for each TM learner associated with the arm. 

\emph{\ac{TM} with Thompson sampling:} We now show how Thompson sampling can be achieved with \ac{TM}. Our approach is similar to that presented in the Tree Bootstrap Algorithm in~\cite{practicalcbdt}. Since a \ac{TM} learner is non-parametric, we use bootstrapping to simulate the behavior of sampling from a posterior distribution. At each time instant $t$, $N$ context reward pairs are bootstrapped from the $D_{t,u}$ with replacement, where $D_{t,u}$ represents the set of observations (context reward pairs) for arm $u$ and $N=|D_{t,u}|$. A \ac{TM} learner is fitted to each of these bootstrapped datasets and at each time $t$, the bootstrapping algorithm selects the arm $u_t$ that has the maximum probability of success $\hat p$.
 For ease of exposition, we outline Thompson Sampling with bootstrapping for \acp{TM} in  Algorithm~\ref{alg:TMthompsonsampling}.

\begin{algorithm}[ht]
   \caption{Thompson Sampling with \ac{TM}}
   \label{alg:TMthompsonsampling}
\begin{algorithmic}
\FOR{$t=1$ {\bfseries} to $T$}
    \STATE get context $s_t$
    \FOR{{\bfseries} $u=1,\dots,K$ }
    \STATE Sample bootstrapped dataset $\tilde D_{t,u}$ of size $N$ from $D_{t,u}$ with replacement.
    \STATE Fit Tsetlin Machine $\mathit{TM}_{t,u}$ to $\tilde D_{t,u}$
    \ENDFOR
\STATE Choose action $u_t=\displaystyle \argmax_{u}(\hat p(\mathit{TM}_{t,u},s_t))$
\STATE Update $D_{t,u_t}$ with $(s_t,r_{t,u_t})$
\ENDFOR
\end{algorithmic}
\end{algorithm}



\section{Empirical Analysis}
\label{sec:numerical}
Given the interactive aspect of contextual bandit algorithms, it is often difficult to evaluate them on real datasets except for a handful number of tasks~\cite{bietti2021contextual}. Therefore, we use standard supervised learning classification datasets for our evaluation. We consider both binary as well as multiclass classification datasets. Each unique label in these datasets is considered as an arm of the equivalent contextual bandit problem, where a separate learner associated with each arm is trained independently.
Our numerical analysis contrasts the performance of \ac{TM} with $\varepsilon-$greedy arm selection and \ac{TM} with Thompson sampling against Tree Bootstrap~\cite{practicalcbdt}, Linear UCB~\cite{chu2011contextual}, Neutral Network ($\varepsilon-$greedy) and Logistic regression ($\varepsilon-$greedy). For our analysis we consider two scenarios described as follows:
\par
\emph{\textbf{Scenario 1: }} In this scenario, we consider five standard supervised learning classification datasets from UCI machine learning repository~\cite{Dua:2019}: \emph{Iris, Breast Cancer Wisconsin (Diagnostic), Adult, Covertype}, and \emph{Statlog (Shuttle)}. The response variables for each of these datasets are, \emph{Iris types, Diagnosis, Occupation}, and \emph{Covertypes}, respectively, where as for the \emph{Statlog(Shuttle)} dataset, we consider the last column as the response variable. Additionally we considered two other classification datasets: \emph{Mnist}~\cite{deng2012mnist} and \emph{Noisy XOR}, where the respective response variables are \emph{digits} and the \emph{XOR output}. The noisy XOR dataset contains the XOR operation of $12$ bit input, 
where the output is flipped with probability $0.4$. These classification
 datasets are converted into a contextual bandit problem where the learner receives a reward of $+1$ for correctly identifying the target value and a reward of $0$ for incorrect classification. The datasets are processed by removing entries with missing values. 

\emph{\textbf{Scenario 2:}} In this scenario, we consider two datasets: \emph{Movielens 100 K}~\cite{harper2015movielens} and \emph{Simulated Article}~\cite{rao2020contextual}. Both these datasets simulate a recommender system. The reward function for the Movielens dataset is the user's rating for a particular movie while for Simulated Article dataset, a reward of $+1$ is received if the recommended article is clicked by the user.  The Movielens dataset is preprocessed, so that top 10 rated movies are selected. One important takeaway from these two datasets is that only the reward function for correct recommendation is provided. For instance the Simulated Article dataset provides recommended articles that has been clicked by the user. Similarly the Movielens dataset provides user movie ratings, where the users only rated the movies they have watched. In such circumstances, the rewards are partial and sparse since there are no information on the ratings of all movies or user click information for each article. To circumvent this, we perform singular value decomposition (SVD) for both the datasets which is a popular approach in collaborative filtering~\cite{koren2009matrix}. For the Movielens dataset, let $S_{i,j}$ be the rating of user $i$ on movie $j$. Low rank SVD of this matrix results in two matrices $W$ and $X$, where $S\approx W*X$. The $i^th$ row of $W$ represents the context features for each user $i$ and the rows of $X$ represents the actions or the movie to be recommended. The reward for recommending a movie $j$ to user $i$ is then the dot product of the corresponding row of $W_i$ and $X_j$. We perform SVD for Simulated Article dataset in a similar manner. In order to obtain binarized  rewards for the actions, the maximum reward corresponding to an arm that yields a reward of $1$, and the rest a reward of $0$. 
For our analysis we consider rank $10$ for both the Movielens and the Simulated Article dataset.  

For both the scenarios we use appropriate binarization before fitting the \ac{TM} learners. The \emph{Noisy XOR} dataset was already binarized and for \emph{MNIST}, we binarize the features having value $\geq 75$ to be $1$ and $0$ otherwise. For other datasets, the maximum number of bits for binarization per feature is given by Table~\ref{tbl:max_bits}.

\begin{table}[!htb]
\resizebox{\textwidth}{!}{%
\caption{Maximum number of bits per feature}
\label{tbl:max_bits}
\begin{tabular}{|l|l|l|}

\hline
\textbf{Datasets}                             & \textbf{Context dim} & \textbf{Max bits per feature }\\ \hline
Iris                                 & 4                 & 4                        \\ \hline
Breast Cancer (Diagnostic) & 30                & 10                       \\ \hline
Adult                                & 15                & 10                       \\ \hline
Statlog Shuttle                      & 9                 & 10                       \\ \hline
Covertype                               & 54              & 10                       \\ \hline
Simulated Article                    & 4                 & 10                       \\ \hline
Movie lens                           & 10                & 8                        \\ \hline
\end{tabular}
}
\end{table}

\begin{table}[!htb]
\resizebox{\textwidth}{!}{%
\caption{\ac{TM} learner configuration}
\label{tbl:TM_config}
\begin{tabular}{|l|l|l|l|l|}
\hline
\textbf{Datasets}                            &\textbf{\#Clauses} & \textbf{T} & \textbf{s}   &\textbf{State bits} \\ \hline
Iris                                 & 1200              & 1000          & 8.0 & 10                   \\ \hline
Breast Cance (Diagnostic) & 650               & 300           & 5.0 & 10                   \\ \hline
Noisy XOR                            & 1000              & 700           & 5.0 & 8                    \\ \hline
Adult                                & 1200              & 800           & 5.0 & 8                    \\ \hline
Statlog Shuttle                      & 1200              & 800           & 5.0 & 8                    \\ \hline
Covertype                               & 1200               & 800           & 5.0 & 8                    \\ \hline
Simulated Article                    & 2000              & 1500          & 5.0 & 10                   \\ \hline
Movie lens                           & 4000              & 3000          & 8.0 & 8                    \\ \hline
MNIST                                & 5000              & 4000          & 5.0 & 8                    \\ \hline
\end{tabular}
}
\end{table}

Table~\ref{tbl:TM_config} shows the configuration of the contextual bandit learners using \ac{TM}. The same \ac{TM}  configurations are used for both \ac{TM} with $\varepsilon-$greedy arm selection and \ac{TM} with Thompson sampling. 

\begin{figure*}[!htb]
\centering %
\floatsetup{floatrowsep=quad}
\begin{floatrow}[3]%
\ffigbox[\FBwidth]{\caption{Iris}\label{fig:figure1}}{\includegraphics[width=0.33\textwidth]{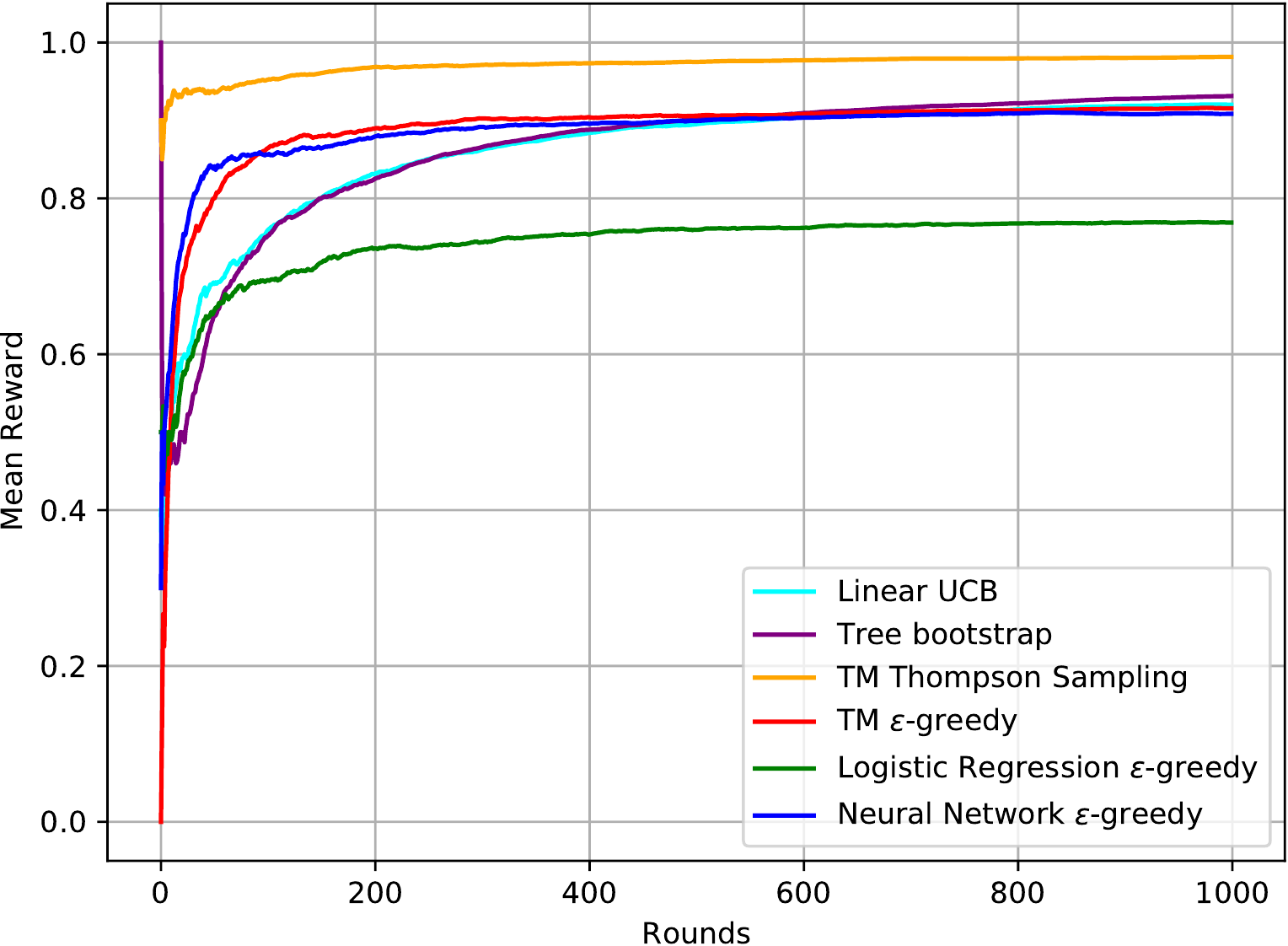}}
\ffigbox[\FBwidth]{\caption{Breast cancer.}\label{fig:figure2}}{\includegraphics[width=0.33\textwidth]{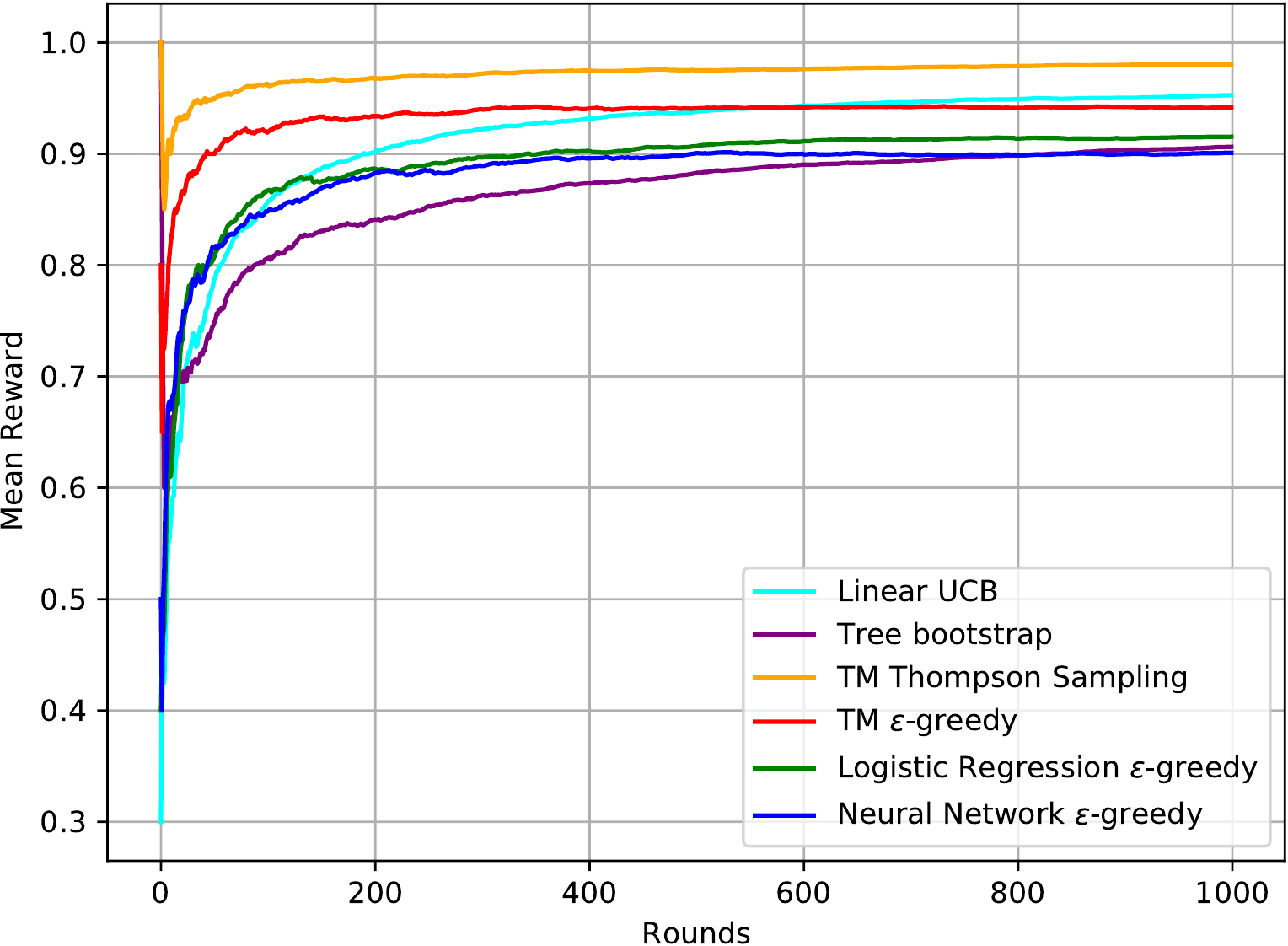}}
\ffigbox[\FBwidth]{\caption{Adult}\label{fig:figure3}}{\includegraphics[width=0.33\textwidth]{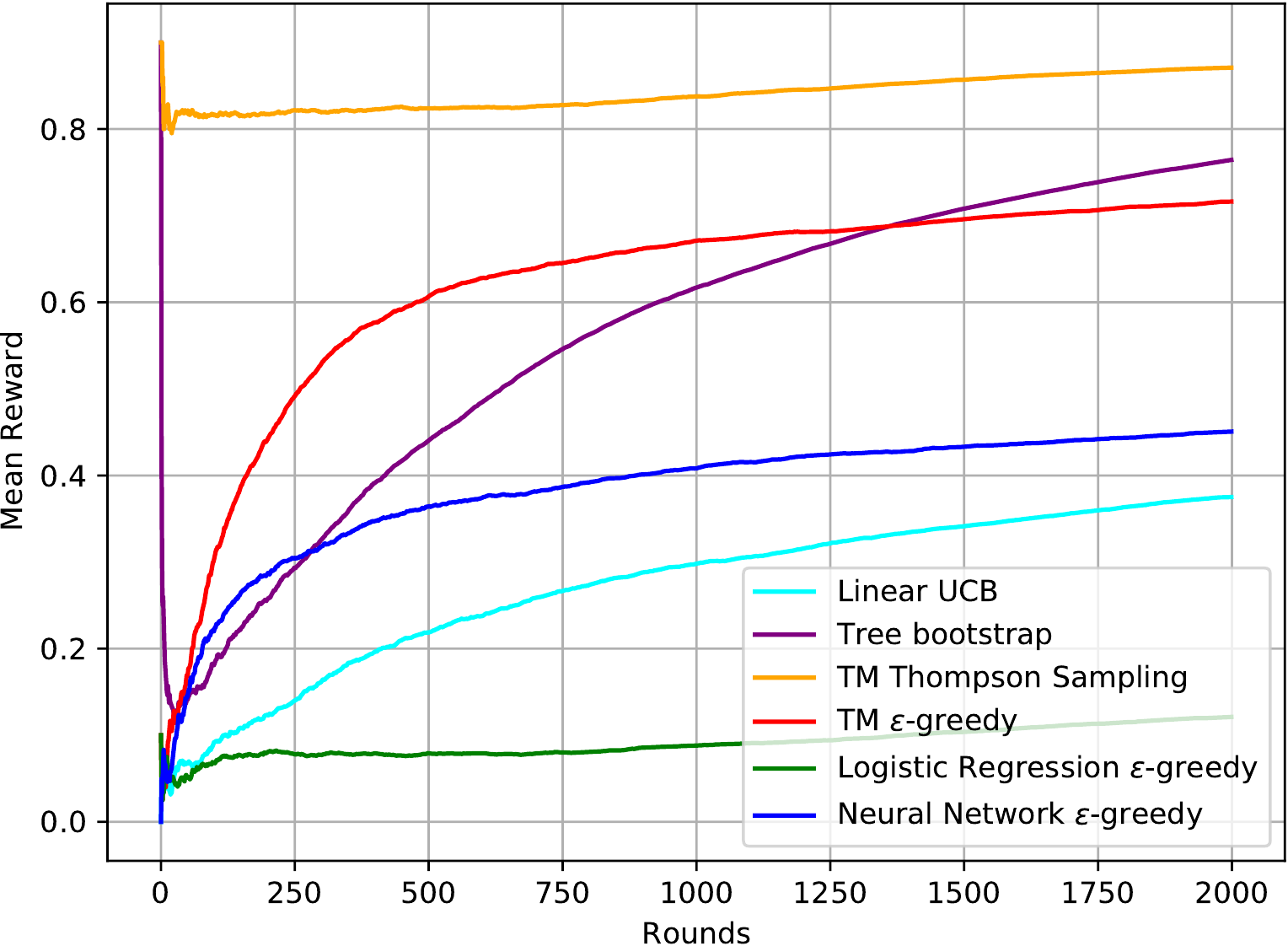}}
\end{floatrow}
\end{figure*}

\begin{figure*}[!htb]
\centering %
\floatsetup{floatrowsep=quad}
\begin{floatrow}[3]%
\ffigbox[\FBwidth]{\caption{Statlog (Shuttle)}\label{fig:figure4}}{\includegraphics[width=0.33\textwidth]{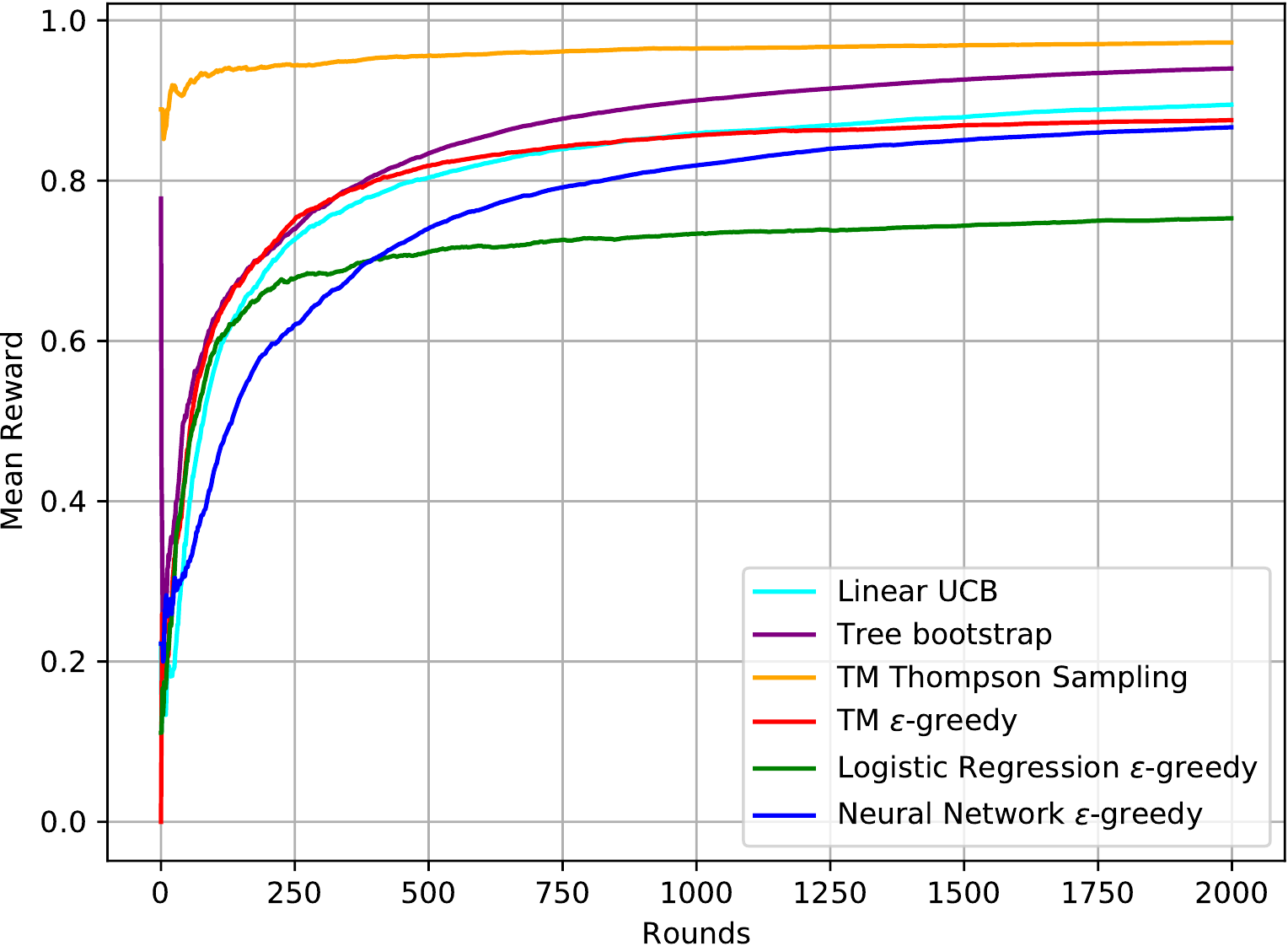}}
\ffigbox[\FBwidth]{\caption{MNIST}\label{fig:figure5}}{\includegraphics[width=0.33\textwidth]{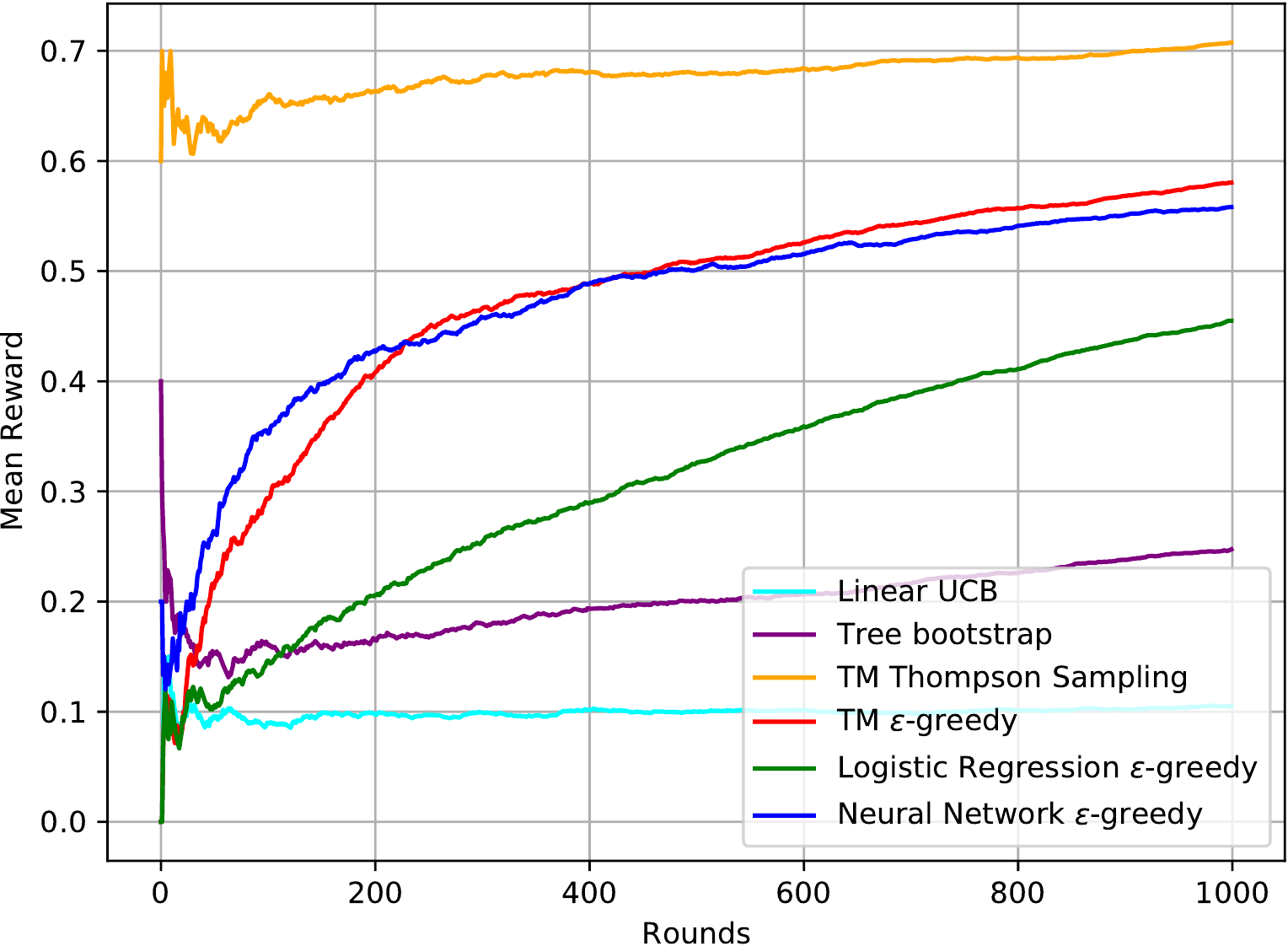}}
\ffigbox[\FBwidth]{\caption{Covertype}\label{fig:figure6}}{\includegraphics[width=0.33\textwidth]{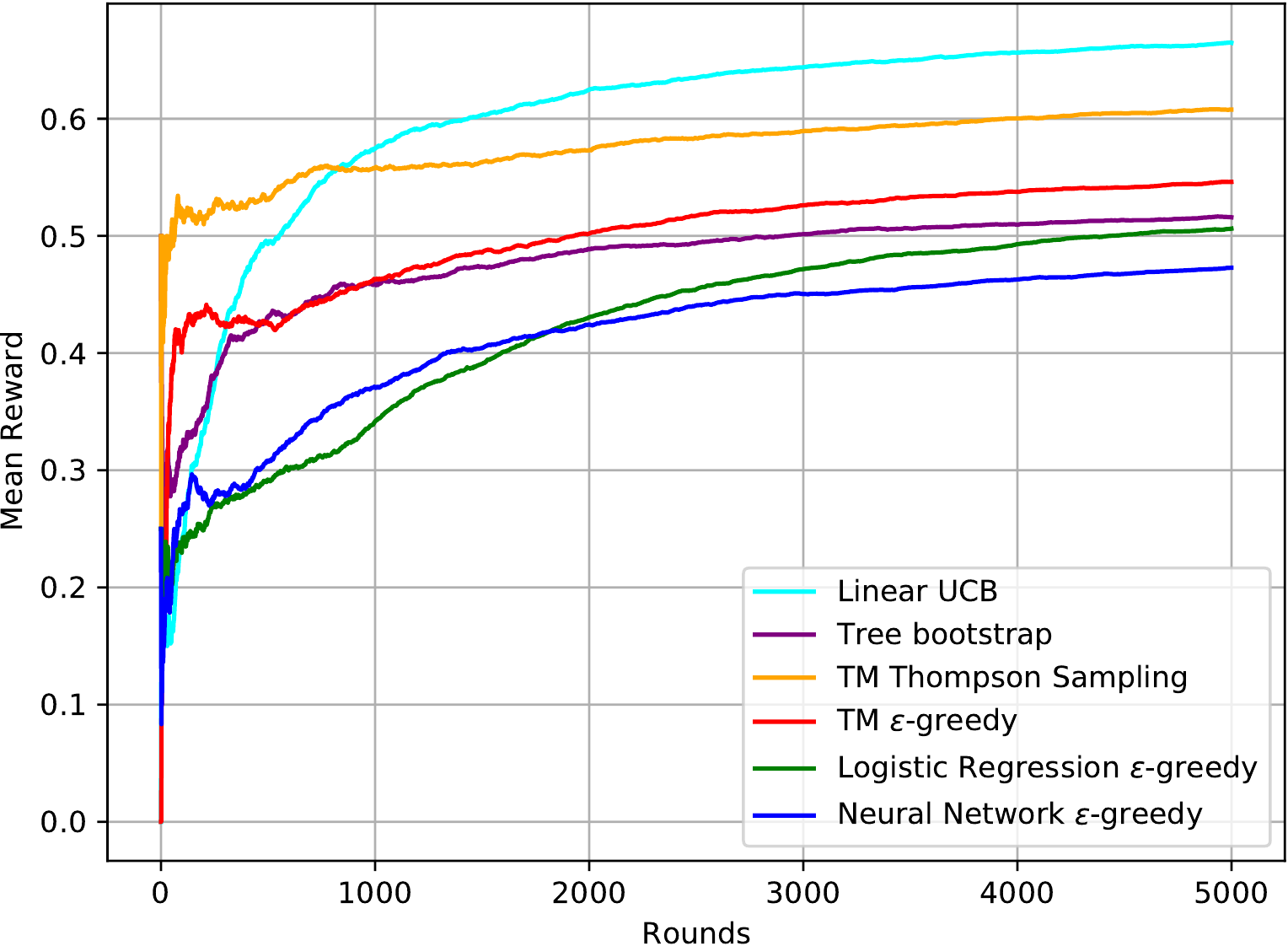}}
\end{floatrow}
\end{figure*}

\begin{figure*}[!htb]
\centering %
\floatsetup{floatrowsep=quad}
\begin{floatrow}[3]%
\ffigbox[\FBwidth]{\caption{Noisy XOR}\label{fig:figure7}}{\includegraphics[width=0.33\textwidth]{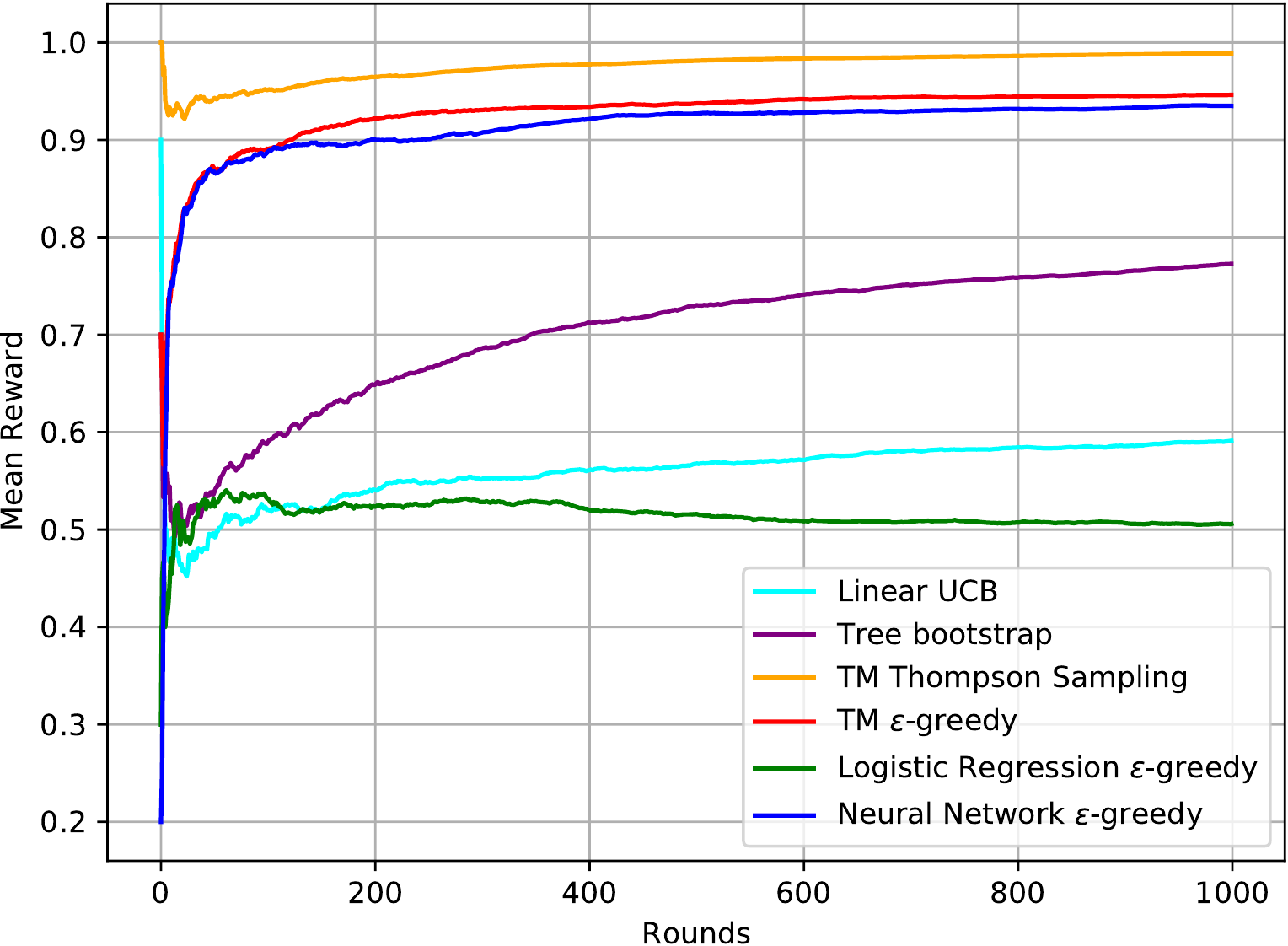}}
\ffigbox[\FBwidth]{\caption{Movielens}\label{fig:figure8}}{\includegraphics[width=0.33\textwidth]{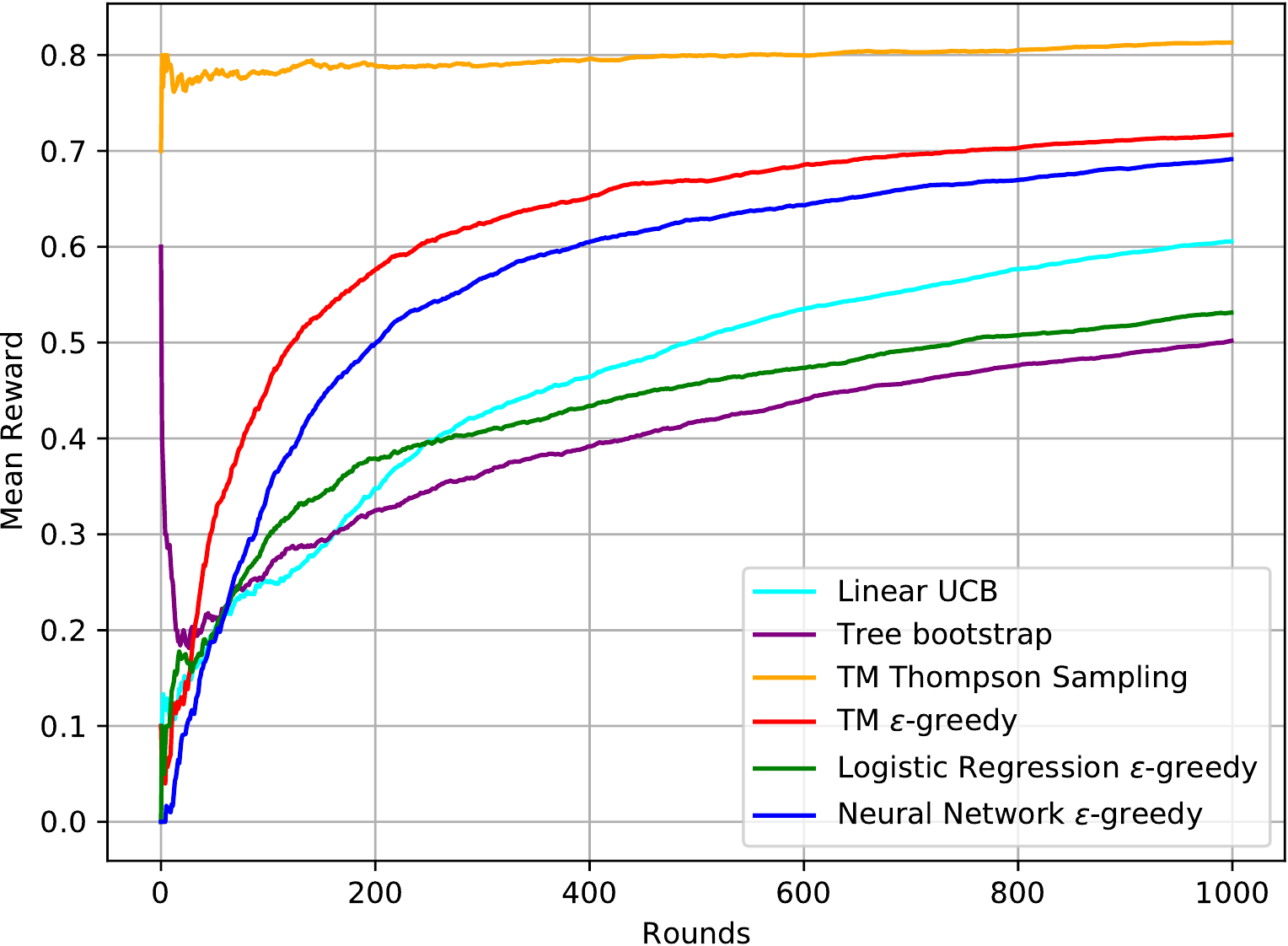}}
\ffigbox[\FBwidth]{\caption{Simulated Article}\label{fig:figure9}}{\includegraphics[width=0.33\textwidth]{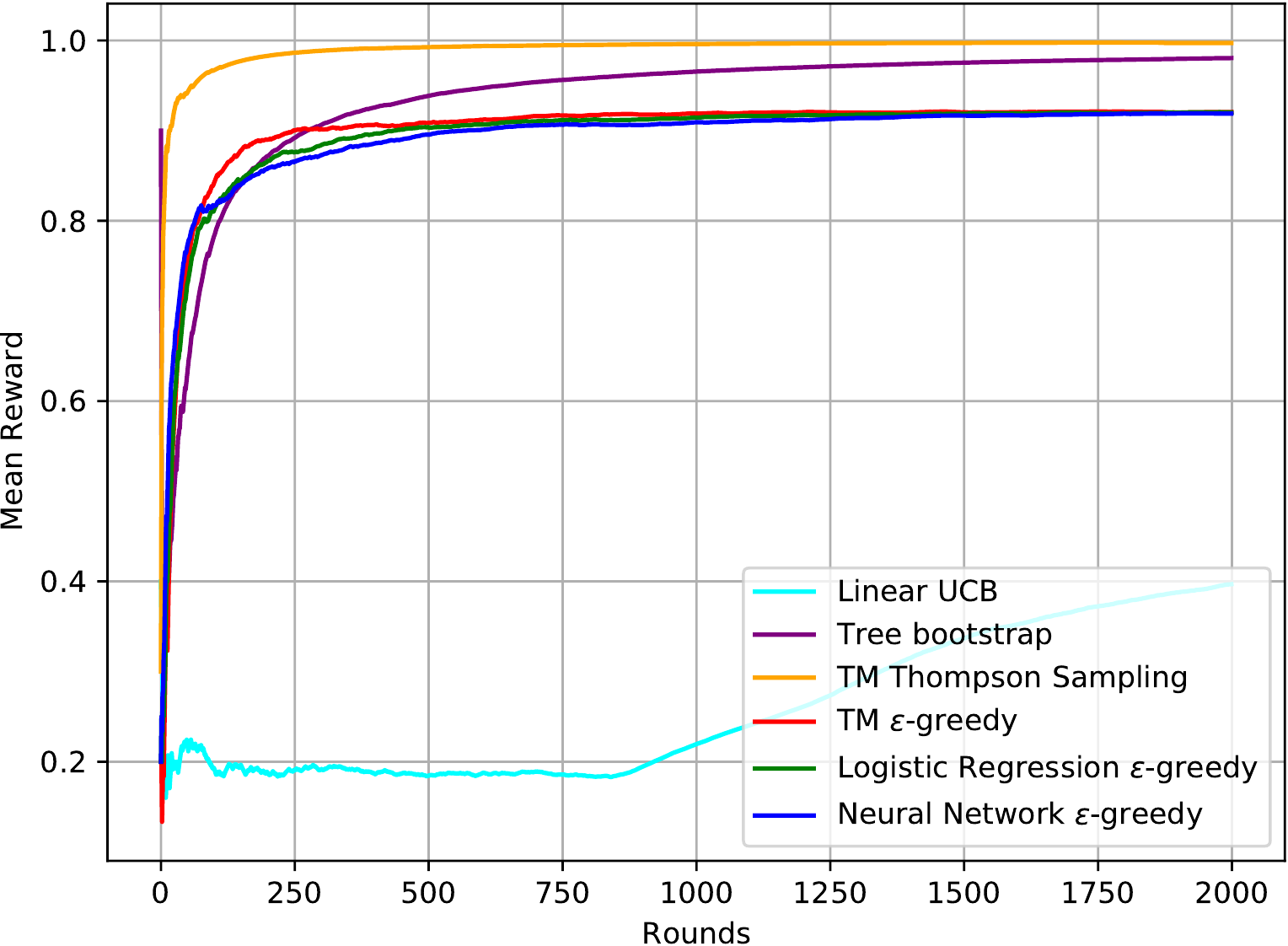}}
\end{floatrow}
\end{figure*}

The results for both the scenarios are presented in Figures~\ref{fig:figure1}-~\ref{fig:figure9}. These results show that for eight datasets, \ac{TM} with Thompson sampling outperforms all other algorithms except for Linear UCB on Covertype. Further, \ac{TM} with $\varepsilon-$greedy provides competitive performance when compared with popular contextual bandit algorithms in the literature. Also,  the \ac{TM} learner learns at a faster rate compared to other learners. These experiments were performed with 10 independent runs for each dataset. The average result across the independent runs are reported. 


\section{Interpretability}
\label{sec:interpretability}
Unlike neural networks and some other complex machine learning approaches, one of the advantages of \acp{TM} is that they produce propositional logic expressions. These are in flat AND-form, which have proven to be human interpretable \cite{human-reasoning}. As explained in Section \ref{sec:TM}, each propositional expression is a conjunctive clause, consisting of features, in their original or negated forms, interacting with each other using logical \emph{AND} operations. These clauses can form a simplified representation of the arm selection policy  by combining them into a single Disjunctive Normal Form (DNF) expression. Since clauses are assigned to each arm of the multi-armed contextual bandit problem, we produce a single DNF expression for each arm. These DNF expressions are propositional logic expressions made up of the context. The \ac{TM} is able to produce these interpretations demonstrating how it interprets the context with respect to each arm.\\
Here, we show the simplified propositional expressions for each arm, obtained from \acp{TM} trained with Thompson Sampling on the Iris dataset:
\begin{enumerate}
    \item[\emph{\textbf{Arm-1:}}] $x_{10} \lor x_{14} \lor x_{15} \lor x_3$
    \item[\emph{\textbf{Arm-2:}}]  $\neg x_1 \lor x_{12} \lor \neg x_{13} \lor \neg x_{14} \lor x_{16} \lor x_8 \lor \neg x_9 \lor (x_{10} \land x_{11} \land x_{15} \land x_2 \land \neg x_4 \land x_5 \land x_6 \land \neg x_7) \lor (x_{10} \land x_{11} \land x_{15} \land x_3 \land \neg x_4 \land x_5 \land \neg x_7)$ 
  \item[\emph{\textbf{Arm-3:}}]  $~\neg x_{10} \lor \neg x_{11} \lor \neg x_{15} \lor (x_1 \land \neg x_{12} \land x_{13} \land x_{14} \land \neg x_{16} \land x_2 \land x_3 \land x_5 \land \neg x_8 \land x_9) \lor (\neg x_{12} \land \neg x_{16} \land x_4)$
\end{enumerate}
The next set of propositional expressions represent the arms for the Simulated Article dataset:
\begin{enumerate}
    \item[\emph{\textbf{Arm-1:}}] 

    $\neg x_4 \lor (x_1 \land \neg x_{10} \land \neg x_{11} \land x_{12} \land x_{13} \land x_{14} \land x_{15} \land x_{16} \land x_{17} \land x_{18} \land x_{19} \land x_2 \land x_{20} \land \neg x_{21} \land x_3 \land x_5 \land x_6 \land x_7 \land x_8 \land x_9)$
  \item[\emph{\textbf{Arm-2:}}]  
  
  
  $(x_1 \land \neg x_{10} \land \neg x_{11} \land x_{12} \land x_{13} \land x_{14} \land x_{15} \land x_{16} \land x_{17} \land x_{18} \land x_{19} \land x_2 \land x_{20} \land \neg x_{21} \land x_3 \land x_4 \land x_5 \land x_6 \land x_7 \land x_8 \land x_9) \lor (\neg x_1 \land x_{10} \land x_{11} \land x_{12} \land x_{13} \land x_{14} \land x_{15} \land x_{16} \land x_{17} \land x_{18} \land x_{19} \land x_2 \land \neg x_{20} \land \neg x_{21} \land x_3 \land x_4 \land x_5 \land x_6 \land \neg x_7 \land x_8 \land x_9)$
  \item[\emph{\textbf{Arm-3:}}]  
  
  $(x_1 \land \neg x_{10} \land \neg x_{11} \land x_{12} \land x_{13} \land x_{14} \land x_{15} \land x_{16} \land x_{17} \land x_{19} \land x_2 \land x_{20} \land \neg x_{21} \land x_3 \land x_4 \land x_5 \land x_6 \land x_7 \land x_9) \lor (\neg x_{11} \land x_{12} \land x_{13} \land x_{14} \land x_{16} \land x_{17} \land x_{18} \land x_{19} \land x_2 \land x_{20} \land \neg x_{21} \land x_3 \land x_4 \land x_5 \land x_8 \land x_9)$
  
  \item[\emph{\textbf{Arm-4:}}]  $x_{10} \land x_{11} \land x_{12} \land x_{13} \land x_{14} \land x_{15} \land x_{16} \land x_{17} \land x_{18} \land x_{19} \land x_2 \land x_3 \land x_4 \land x_5 \land x_6 \land x_8 \land x_9 \land \neg x_1 \land \neg x_{20} \land \neg x_{21} \land \neg x_7$
\end{enumerate}
The propositional expressions shown above exhibit how the \ac{TM} learns interactions between propositional input features (literals $x_i$). As seen, the interactions are learned with logical \emph{AND} operations between original and negated features. These learnt interactions (clauses) are combined with the logical \emph{OR} operation (or by addition for $T > 1$. From the propositional expressions, we can see which arm is selected by \ac{TM} by simply plugging in the input values. An arm is selected if its propositional expression evaluates to True (or it obtains the largest net sum for $T > 1$). This is arguably an important advantage of \acp{TM} --- a trained \ac{TM} model can be reduced down to simple propositional expressions. Of course, as such, any machine learning algorithm can be converted into logical form. However, the complexity of such representations can be immense for the competitive classic machine learning models, which the \ac{TM} still outperforms by a large margin \cite{DC}.

\section{Conclusion}
\label{sec:conclusion}
In this paper, we presented an interpretable and practical contextual bandit learner using Tsetlin Machine (TM). Our analysis showed that \ac{TM} as a contextual bandit learner provides competitive performance compared to other popular contextual bandit algorithms. We then presented how Thompson sampling can be implemented using \ac{TM}, where our approach is derived from the Tree Bootstrap algorithm. Finally, we perform interpretability analysis where the arm selection strategy can be characterized by a propositional function of the contexts. One of the limitations of using \ac{TM} is that it requires the contexts to be binarized. While such binarization leads to a loss of information, our empirical analysis shows that \ac{TM} with Thompson sampling performs substantially better than the other evaluated learners on the majority of the datasets. Having such promising empirical results, we aim at providing theoretical performance guarantees for algorithms with \ac{TM} as future work.

\bibliography{references}
\bibliographystyle{icml2022}



\end{document}